\documentclass[conference]{IEEEtran}
\IEEEoverridecommandlockouts
\usepackage{cite}
\usepackage{amsmath,amssymb,amsfonts}
\usepackage{graphicx}
\usepackage{textcomp}
\usepackage{xcolor}
\def\BibTeX{{\rm B\kern-.05em{\sc i\kern-.025em b}\kern-.08em
    T\kern-.1667em\lower.7ex\hbox{E}\kern-.125emX}}

\usepackage{algorithm}
\usepackage{algorithmicx}
\usepackage[noend]{algpseudocode}
\usepackage[caption=false,font=normalsize,labelfont=sf,textfont=sf]{subfig}
\usepackage{amsmath}
\DeclareMathOperator*{\argmax}{argmax}
\usepackage[absolute]{textpos}

\begin{document}

\begin{textblock}{12}(2,0.3)
	\noindent Please cite as follows:
	K. Malialis, C. G. Panayiotou and M. M. Polycarpou, ``Data-efficient Online Classification with Siamese Networks and Active Learning'', 2020 International Joint Conference on Neural Networks (IJCNN), Glasgow, United Kingdom, 2020, doi: 10.1109/IJCNN48605.2020.9206730.
\end{textblock}

\title{Data-efficient Online Classification with\\Siamese Networks and Active Learning}


\author{
	\IEEEauthorblockN{
		Kleanthis Malialis\textsuperscript{a},
		Christos G. Panayiotou\textsuperscript{a, b} and
		Marios M. Polycarpou\textsuperscript{a, b}
	}
	\IEEEauthorblockA{
		\textsuperscript{a} \textit{KIOS Research and Innovation Center of Excellence}\\
		\textsuperscript{b} \textit{Department of Electrical and Computer Engineering}\\
		\textit{University of Cyprus}\\
		Nicosia, Cyprus\\
		Email: \{malialis.kleanthis, christosp, mpolycar\}@ucy.ac.cy\\
		ORCID: \{0000-0003-3432-7434, 0000-0002-6476-9025, 0000-0001-6495-9171\}
	}
}

\maketitle

\begin{abstract}
An ever increasing volume of data is nowadays becoming available in a streaming manner in many application areas, such as, in critical infrastructure systems, finance and banking, security and crime and web analytics. To meet this new demand, predictive models need to be built online where learning occurs on-the-fly. Online learning poses important challenges that affect the deployment of online classification systems to real-life problems. In this paper we investigate learning from limited labelled, nonstationary and imbalanced data in online classification. We propose a learning method that synergistically combines siamese neural networks and active learning. The proposed method uses a multi-sliding window approach to store data, and maintains separate and balanced queues for each class. Our study shows that the proposed method is robust to data nonstationarity and imbalance, and significantly outperforms baselines and state-of-the-art algorithms in terms of both learning speed and performance. Importantly, it is effective even when only $1\%$ of the labels of the arriving instances are available.
\end{abstract}


\section{Introduction}\label{sec:intro}
Traditionally, predictive models are built from historical data consisting of examples annotated with class labels (i.e. the ground truth). This paper is concerned with online learning, with a focus on the following key challenges:
\begin{itemize}
	\item \textbf{One-by-one online learning}: We focus on online learning i.e. as data is arriving in a streaming fashion. Contrary to the majority of online learning work, we focus on one-by-one learning, where only a single instance (rather than a batch) is observed at each time step.
	\item \textbf{Limited labelled data}: Acquiring labels at every step is expensive / impractical. A potential solution that makes such an assumption, may not be practical in real applications. In this paper, we consider limited labelled data.
	\item \textbf{Nonstationary data}: This paper focuses on online learning where the distribution of data is unknown. It is also concerned with cases where the data distribution is nonstationary; i.e., it evolves or ``drifts'' over time.
	\item \textbf{Imbalanced data}: Class imbalance, in conjunction with the aforementioned challenges, causes one-by-one online learning to become significantly more challenging.
\end{itemize}

The contributions of this paper are as follows. We provide new insights into learning from limited labelled, nonstationary and imbalanced in one-by-one online classification, a largely unexplored area. We propose a novel learning approach for one-by-one online learning which utilises active learning and siamese neural networks. Active learning is a paradigm in which the classifier selectively queries an oracle (typically, a human expert) to provide class labels according to an allocated budget \cite{settles2009active}. Several industrial large-scale classification systems, such as, Google's method for labeling malicious advertisements, have been realised through active learning \cite{sculley2011detecting}.

Siamese networks enable learning when only a few examples per class are available, commonly referred to as \textit{few-shot learning}, and have recently achieved state-of-the-art results in image recognition \cite{koch2015siamese}. To our knowledge, this is the first time that an approach is developed, which synergistically brings together siamese networks and one-by-one active learning. The proposed synergy enables the effective learning from limited labelled data in nonstationary and imbalanced settings.

The organisation of this paper is as follows. Section~\ref{sec:background} provides the background material necessary to understand the contributions made in the paper. Section~\ref{sec:related_work} provides a review of related work. The proposed learning approach is presented in Section~\ref{sec:proposed}. Our experimental setup is described in Section~\ref{sec:exp_setup} while the experimental results are presented and discussed in Section~\ref{sec:exp_results}. Lastly, concluding remarks and directions for future work are provided in Section~\ref{sec:remarks}.

\section{Background}\label{sec:background}
In \textbf{online} learning we consider a data generating process that provides at each time step $t$ a sequence of examples or instances $S^t = \{(x^t_i,y^t_i)\}^M_{i=1}$ from an unknown probability distribution $p^{t}(x,y)$, where $x^t \in \mathbb{R}^d$ is a $d$-dimensional input vector belonging to input space $X \subset \mathbb{R}^d$, $y^t \in [1, K]$ is the class label, $K \geq 2$ is the number of classes and $M$ is the number of instances arriving at each step.

When the observed sequence $S^t$ consists only of a single instance (i.e. $M=1$), it is termed \textbf{one-by-one online} learning, otherwise it is termed \textbf{batch-by-batch online} learning \cite{ditzler2015learning}. The design of batch-by-batch algorithms differs significantly from that of one-by-one algorithms as they are designed to process chunks of data, possibly by utilising an offline learning algorithm \cite{wang2018systematic}. Therefore, the majority of batch-by-batch algorithms are typically not suitable for one-by-one tasks \cite{wang2018systematic}. This work focuses on one-by-one online learning, which is important for real-time monitoring and control.

\textbf{Active} learning is concerned with strategies to selectively query for labels from an \textit{oracle} (typically, a human expert) according to a set of available resources \cite{settles2009active}. Typically, the available resources are modelled by an allocated budget $B \in [0,1]$ where it is expressed as a fraction of the number of arriving examples e.g. $B=0.1$ means that $10\%$ of the arriving instances can be labelled \cite{zliobaite2013active}. A budget spending mechanism must ensure that the labelling spending $b \in [0,1]$ does not exceed the allocated budget.

In one-by-one online classification, a classifier is built that receives a new example $x^t$ at time $t$ and makes a prediction $\hat{y}^t$ based on a concept $h: X \to Y$ such that $\hat{y}^t = h(x^t)$. A given active learning strategy $\alpha : X \to \{0,1\}$ determines if the true label $y^t$ is required, which is assumed that the oracle will provide. The classifier is evaluated using a loss function and is then trained, i.e., its parameters are updated accordingly based on the loss incurred. This process is repeated at each step and, depending on the application, new examples do not necessarily arrive at regular and pre-defined intervals. If learning occurs on the most recent single instance (or batch) only, without taking into account previously labelled data, it is termed \textbf{incremental} (or \textit{one-pass}) learning \cite{ditzler2015learning}. Specifically, the cost $J$ at time $t$ is calculated using the loss function $l$ as follows $J=l(y^t,\hat{y}^t)$.

Learning in \textbf{nonstationary environments} is a major challenge in some applications. Nonstationarity is caused by \textbf{concept drift}, which represents a change in the joint probability. Drift can be characterised by type, severity, speed, predictability, frequency and recurrence \cite{minku2010impact}. Hence, in practise, it is very difficult to characterise concept drift. Our focus is on \textit{learning the concept drift} without its explicit characterisation and detection. \textbf{Class imbalance} \cite{he2008learning} is another challenge that occurs when at least one data class is under-represented compared to others, thus constituting a minority class.

\section{Related Work}\label{sec:related_work} 
Algorithms that are capable of learning from imbalanced and nonstationary data in one-by-one online classification typically fall into two categories: \textit{(i)} resampling algorithms (e.g. \textit{QBR} \cite{malialis2018queue}, \textit{OOB} \cite{wang2015resampling}) and \textit{(ii)} cost-sensitive learning algorithms (e.g. \textit{CSOGD} \cite{wang2014cost}). These have been shown to perform well provided that the data label becomes available at each time step. This may be a key limitation in some applications since acquiring data labels is expensive or impractical to do at every single time step. This work focuses on active learning to address this problem. We provide a description of existing active learning strategies and budget spending mechanisms.

\subsection{Active learning strategies}\label{sec:al_strategies}
The most common active learning strategy is \textbf{uncertainty sampling}, where the learner queries the most uncertain instances, which are typically found around the decision boundary. One way to measure uncertainty is to query the instance whose best labelling is the least confident \cite{settles2009active}. The majority of active learning work assumes the availability of all training examples (offline active learning) \cite{cohn1994improving} while some work considers batch-by-batch online active learning \cite{zhu2007active}.

Recently the community started focusing on one-by-one active learning \cite{zliobaite2013active}. The arriving $x^t$ is queried if it satisfies:
\begin{equation}
p(y^* | x^t) < \theta,
\end{equation}
\noindent where $y^* = \argmax_y p(y|x^t)$ and $\theta$ is a threshold which is typically fixed. This is known as a \textit{fixed uncertainty} strategy

In \cite{zliobaite2013active} the authors introduce a \textit{randomised variable uncertainty} strategy. A fixed uncertainty strategy may fail if the threshold is set incorrectly, or if the classifier learns enough so that the uncertainty remains above the fixed threshold most of the time. The threshold is modified as follows:
\begin{equation}\label{eq:strategy}
	\theta =
	\begin{cases}
		\theta (1 - s) & \text{if } p(y^* | x^t) < \theta_{rdm}\\
		\theta (1 + s) & \text{if } p(y^* | x^t) \geq \theta_{rdm}\\
	\end{cases}
\end{equation}
\noindent where $s$ is a step size parameter, $\theta_{rdm} = \theta * \eta \sim N(1,\delta)$ and $\delta$ is another parameter. Randomisation ensures that the probability of labelling an instance is not zero. This strategy has been shown to work very well.

Uncertainty sampling has been criticised as being prone to outliers \cite{settles2009active} and, for this reason, \textbf{density sampling} has been proposed. Its central idea is that informative queries are not only those that are uncertain, but those which lie in high density regions. As with uncertainty sampling, the majority of work is on offline active learning where the set of all unlabelled instances $U$ is already available. In \cite{settles2008analysis} the instance queried for labelling is selected by its average similarity to other instances in $U$ as follows:
\begin{equation}
\argmax_x \frac{1}{U} \sum^U_{u=1} sim(x, x_u),
\end{equation}
where $sim$ is the cosine similarity. Density sampling has also been applied in batch-by-batch online active learning  where, using the Mahalanobis distance, an informative instance is the one which is similar to other unlabelled instances in the most recent batch \cite{capo2013active}. In our work, we use density sampling in one-by-one active learning. The interested reader is directed towards \cite{settles2009active} for a survey on active learning strategies.

\subsection{Budget spending mechanisms}
A budget spending mechanism must ensure that the label spending $b \in [0,1]$ does not exceed the allocated budget $B \in [0,1]$ over infinite time. The most common approach is to count the \textit{exact} labelling spending \cite{zliobaite2013active}. The labelling expenses at any time $t$ is given by:
\begin{equation}
b^t= \frac{u^t}{t},
\end{equation}
\noindent where $u^t$ is the number of instances queried until time $t$. The drawback of this mechanism is that the contribution of every next label will diminish over infinite time.

One way to solve the aforementioned problem is to count the \textit{exact} labelling spending over a sliding window $w$:
\begin{equation}
b^t= \frac{u^t_w}{w},
\end{equation}
\noindent where $u^t_w$ is the number of instances queried within the sliding window. This, however, defies the requirements of incremental learning. The authors in \cite{zliobaite2013active} propose to \textit{approximate} the number of instances queried within the sliding window:
\begin{equation}\label{eq:budget}
\hat{u}^t_w = 	\lambda \hat{u}^{t-1}_w + a(x^t),
\end{equation}
\noindent where $\lambda = \frac{w-1}{w}$. The authors prove that $\hat{b}$ is an unbiased estimate of $b$.

\section{Proposed approach}\label{sec:proposed}
\begin{figure}[t!]
	\centering
	\includegraphics[scale=0.33]{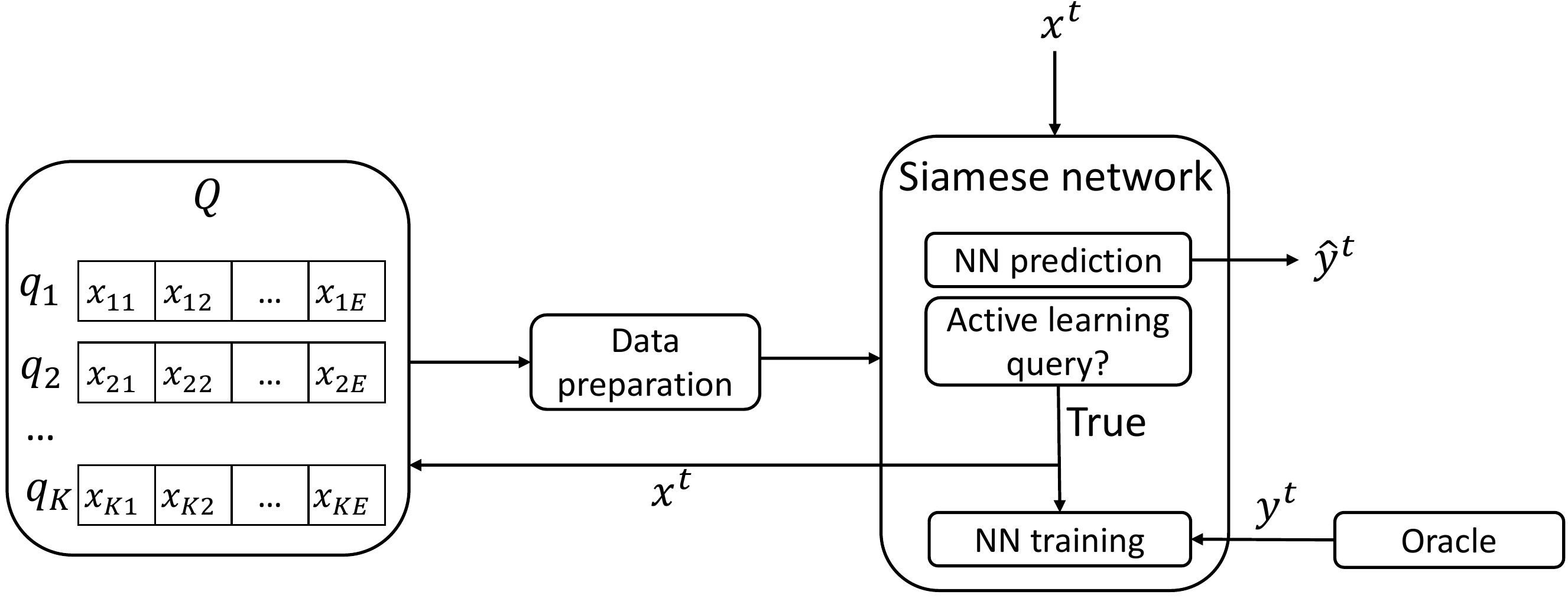}
	
	\caption{Overview of the proposed method which uses a multi-sliding window approach to data storage, a siamese neural network (NN) and active learning to enable online learning from limited, nonstationary and imbalanced data.}
	
	\label{fig:overview}
\end{figure}

The overview of the proposed learning approach is shown in Fig.~\ref{fig:overview}. The learning approach uses a multi-sliding window approach to store data, denoted by $Q$ in the figure. A data preparation phase is in place before feeding the data to a siamese neural network. For each arriving example $x^t$, the siamese network makes a prediction $\hat{y}^t$. Notice that, only if the active learning strategy decides to send a query, the new example is stored in the $Q$ and the siamese network is then trained. A detailed description of each component is provided below, followed by a discussion on the advantages and limitations of the proposed approach.

\subsection{Detailed Description}
\textbf{Data storage}: Our approach assumes the initial availability of $E$ examples per class. Importantly, to avoid a potential deployment issue, we restrict $E$ to be a very small number, e.g., up to five. We argue that for the vast majority of applications this assumption is realistic.

The initial labelled examples and those which will be queried by the active learning strategy will be stored using a multi-sliding window approach. Each sliding window is implemented as a queue and at any time $t$, we maintain a collection of $K$ queues for each class:
\begin{equation}
Q^t = \{q^t_c\}^{K}_{c=1},
\end{equation}
\noindent where $K \geq 2$ is the number of classes. All queues have the same capacity and each queue is defined as follows:
\begin{equation}
q^t_c = \{x_{ci}\}^{E}_{i=1},
\end{equation}
\noindent where for any two $x_{ci}, x_{cj} \in q^t_c$ such that $j > i$, $x_{cj}$ arrived more recently in time. The multi-sliding window approach is depicted by $Q$ in the figure. Notice that we have set the capacity of each queue to $|q^t_c| = E$. The advantage of this is two-fold. Firstly, the number of examples required for storage is small and secondly, the queues will always remain balanced thus avoiding any bias towards majority classes.

\begin{figure}[t!]
	\centering
	\includegraphics[scale=0.28]{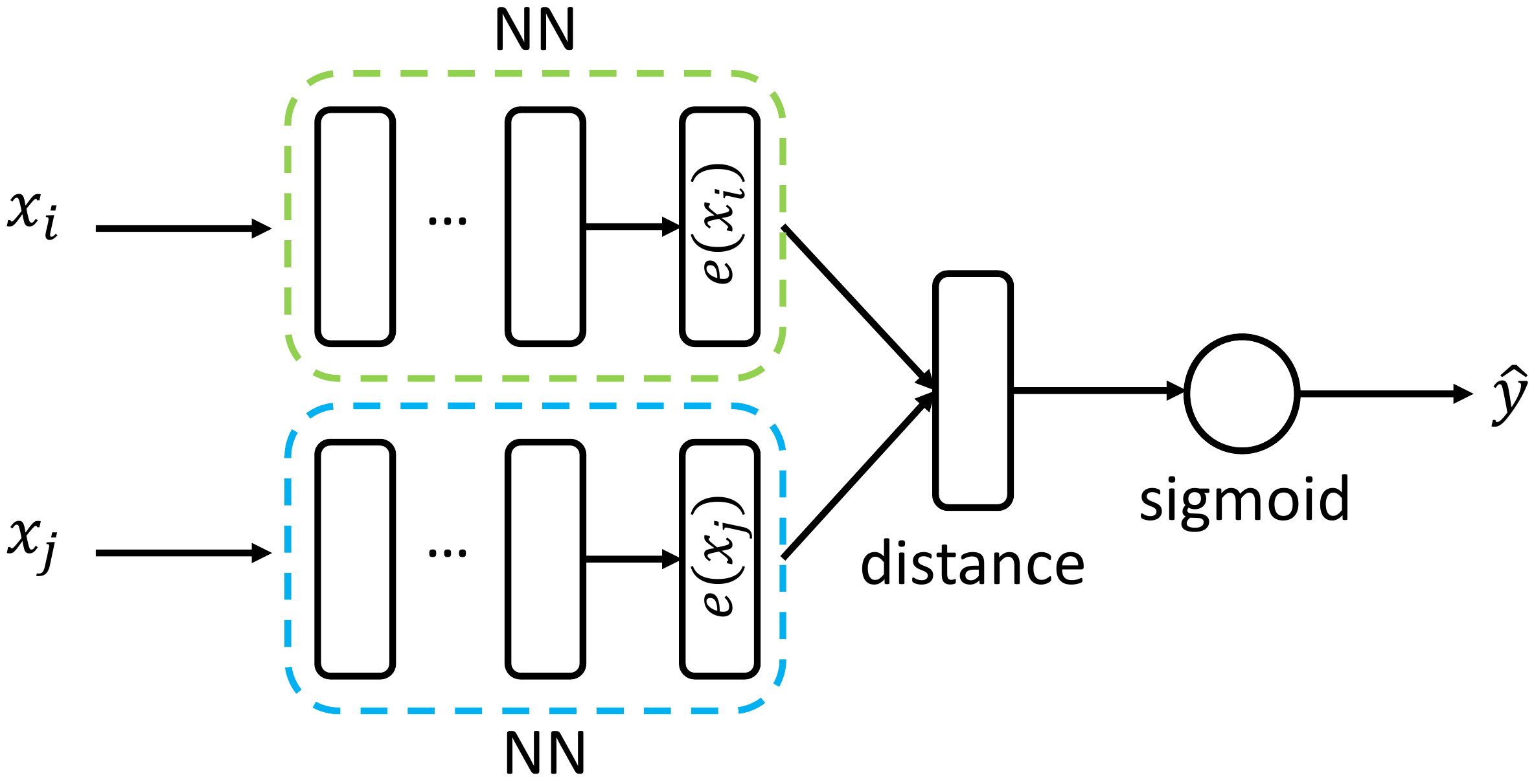}
	
	\caption{A siamese neural network (NN)}
	
	\label{fig:siamese}
\end{figure}

\textbf{Siamese network \& training}: At the heart of the proposed approach lies a siamese neural network \cite{chopra2005learning}. A siamese network consists of two identical neural networks (the `twins') as shown in Fig.~\ref{fig:siamese}. The central idea is to learn a function (denoted by $e$ on the figure) that maps an input pattern into a target space (the `embedding') in such a way that a simple distance in the target space approximates the neighbourhood relationships in the input space. For this reason, siamese networks have been shown to significantly outperform traditional models such as k-nearest neighbour (k-NN) algorithms in high-dimensional spaces e.g. for image recognition \cite{koch2015siamese}. The element-wise absolute difference is considered as the distance metric:
\begin{equation}
d(x_1, x_2) = | e(x_1) - e(x_2) |
\end{equation}

This is then given to a single sigmoidal output unit. Ideally, the siamese network $h: X \times X \rightarrow Y$ will learn to output $\hat{y} = 1$ for any pair of inputs that belongs to the same class and $\hat{y} = 0$ if the pair of inputs is of different class. There exists a process which transforms the examples in $Q^t$ into the training pairs $Q^t_{train}$ (described in the next section). The cost function used is then as follows:
\begin{equation}\label{eq:cost}
J= \frac{1}{|Q^t_{train}|} \sum_{(x_1, x_2) \in Q^t_{train}} l(y^t, h(x_1, x_2))
\end{equation}
\noindent where the loss function $l$ used is the binary cross-entropy.

\textbf{Data preparation}: Given the data in $Q^t$ at time $t$, we generate all possible combinations $C^t_2$ of size two. Three subsets of $C^t_2$ are then generated as follows.

The first one ($Q_{id}$) contains all pairs in which the two examples are identical and, hence, belong to the same class. The second one ($Q_{same}$) contains all (non-identical) pairs in which the two examples belong to the same class. The two are then joined ($Q_{id\_same}^t$) to indicate the positive pairs. These are defined as follows:
\begin{align}
	Q_{id}^t & = \{(x_{c_1, i_1}, x_{c_2, i_2}) \in C^t_2 | c_1 = c_2, i_1 = i_2 \}\\
	Q_{same}^t & = \{(x_{c_1, i_1}, x_{c_2, i_2}) \in C^t_2 | c_1 = c_2, i_1 \neq i_2 \}\\
	Q_{id\_same}^t & = Q_{id}^t \cup Q_{same}^t
\end{align}
It is essential that $Q_{same}^t$ is not empty; in other words, the approach expects that the initial labelled set consists of at least two examples per class i.e. $E \geq 2$.

The third subset ($Q_{diff}$) contains all pairs in which the two examples belong to different queues:
\begin{equation}
Q_{diff}^t = \{(x_{c_1, i_1}, x_{c_2, i_2}) \in C^t_2 | c_1 \neq c_2 \}
\end{equation}

Importantly, resizing is performed to ensure balance between positive and negative pairs. The training set is thus formed as follows:
\begin{equation}\label{eq:training_set}
Q_{train}^t = Q_{id\_same}^t \cup Q_{diff}^t
\end{equation}

\textbf{Class prediction}: The siamese network predicts the class of each arriving instance $x^t$ by taking into consideration all examples in the queues $Q^t$. For each queue, we find the average similarity of $x^t$ to its elements. We then choose the queue with the highest average similarity as follows:
\begin{equation}\label{eq:predict}
\argmax_{c \in [1, K]}  \frac{1}{|E|} \sum_{i=1}^E p(y | x^t, x_{ci}^t)
\end{equation}

\textbf{Active learning strategy}: One of the advantages of this approach is that it is highly flexible with respect to the active learning strategy as it does not rely upon any specific strategy. Since the proposed approach is intended to address one-by-one online classification tasks, however, it is expected that a one-by-one active learning strategy will be used.

This work uses a \textit{randomised variable similarity} strategy, inspired from the \textit{randomised variable uncertainty} strategy. In fact, the equation is the same as the one described in Eq.~\ref{eq:strategy}, although, the selection criterion is not $p(y^* | x^t)$ but the maximum similarity in the predicted class:
\begin{equation}\label{eq:criterion}
\max_i p(y | x^t, x_{ci}^t)
\end{equation}
\noindent where $c$ is the class selected using Eq.~\ref{eq:predict}. The budget spending mechanism used is the one shown in Eq.~\ref{eq:budget}. The pseucode of the proposed learning approach is presented in Algorithm~\ref{alg:proposed}.

\begin{algorithm}[t]
	\caption{Proposed learning method}
	\label{alg:proposed}
	\begin{algorithmic}[1]
		
		\Statex \textbf{Input:}
		\Statex $a$: active learning strategy
		\Statex $B$: labelling budget
		
		\State $Q^0$: initial labelled examples
		\State $h^0$: siamese network
		\State $b^0 = 0$: budget expenses
		
		\For{each time step $t$}
		\State receive example $x^t \in \mathbb{R}^d$
		\State predict class using Eq.\ref{eq:predict}
		\State $h^t = h^{t-1}$
		\State $Q^t = Q^{t-1}$
		
		\If{$b^{t-1} < B$}\Comment expenses within budget
		\State calculate query criterion value using Eq.\ref{eq:criterion}
		\If{$a(x^t, v) == True$}\Comment label request
		\State receive true label $y^t$
		\State append $x^t$ to relevant queue in $Q^t$
		\State prepare training pairs $Q^t_{train}$ using Eq.~\ref{eq:training_set}
		
		\State calculate cost $J$ using Eq.~\ref{eq:cost}
		\State update classifier $h^t = h^{t-1}.train()$
		\EndIf
		\EndIf
		
		\State update budget expenses $b^t$ using Eq.~\ref{eq:budget}
		\EndFor
		
	\end{algorithmic}
\end{algorithm}

\subsection{Discussion}
\textbf{Class imbalance.} The proposed approach is robust to class imbalance. This is the result of three mechanisms `embedded' in the approach. Firstly, the use of \textit{separate} and \textit{balanced} queues alleviates the problem as propagating past examples in the most recent training set can be viewed as a form of oversampling. This concept has been applied in \cite{malialis2018queue} for binary one-by-one online classification tasks. This work extends this to a multi-sliding window approach. Secondly, the data preparation phase creates $|Q^t_{train}|$ training pairs, as opposed to the $K \times E$ examples in the $Q^t$. Depending on the values of $K$ and $E$, the number of training pairs can be considerably larger thus constituting another form of oversampling. Thirdly, we always balance the number of positive and negative pairs. As we will demonstrate in our experimental work, the learning approach can perform well even in extreme imbalanced scenarios.

\textbf{Concept drift.} As it will be illustrated, the learning approach is robust to drift too. As the examples are carried over a series of time steps, this allows the classifier to `remember' old concepts. The classifier needs to also be able to `forget' old concepts. This is achieved by the algorithm's memory-based nature i.e. by bounding the length of queues, these are behaving like sliding windows.

\textbf{Fixed memory.} The proposed learning algorithm is not incremental. An incremental learning algorithm would receive instance $x^t$, its active learning strategy would decide if training will be performed, and then would discard $x^t$. The proposed algorithm, despite not being incremental, always uses a fixed amount of memory that contains $K \times E$ examples. Additionally, the storage requirements are low since the number of examples per class is kept to a minimum; e.g. $|q^t_c| = E \in \{2, 3, 4, 5\}$ Most importantly, however, we will demonstrate that an incremental learning algorithm performs significantly worse compared to algorithms that utilise the examples in $Q^t$.

\section{Experimental Setup}\label{sec:exp_setup}

\subsection{Data}
Synthetic datasets provide us with the flexibility to control various parameters of the approach; e.g., the severity of class imbalance, when to introduce concept drift and the drift characteristics. Synthetic datasets enable us to stress test the proposed approach. We will use the following datasets.

\textbf{sea4} \cite{street2001streaming}: It has two features $x_1,x_2 \in [0,10]$ and four classes. The decision boundaries are as follows:
\begin{equation}
\begin{split}
0 \leq x_1 + x_2 < \theta_1 \longrightarrow class \: 1\\
\theta_1 \leq x_1 + x_2 < \theta_2 \longrightarrow class \: 2\\
\theta_2 \leq x_1 + x_2 < \theta_3 \longrightarrow class \: 3\\
\theta_3 \leq x_1 + x_2 \leq 10 \longrightarrow class \: 4\\
\end{split}
\end{equation}

We choose the thresholds as follows $\theta_1 = 3.0$, $\theta_2 = 5.0$ and $\theta_3 = 7.0$. When concept drift occurs, the thresholds are changed abruptly to  $\theta_1 = 2.0$, $\theta_2 = 6.0$ and $\theta_3 = 8.0$. Data normalisation is afterwards applied so that $x_1, x_2 \in [0, 1]$.

Initially, the dataset is balanced i.e. the probability of an arriving instance belonging to any class is $p(y) = 0.25$. Also, we conducted experiments in a multi-minority scenario where the probability of an arriving instance belonging to a specific class is $p(y) = 0.97$, while for the other three is $p(y) = 0.01$. Notice that this constitutes a case of severe imbalance as our aim is to stress test our learning approach.

\begin{figure}[t!]
	\centering
	
	\subfloat[original]{\includegraphics[scale=0.12]{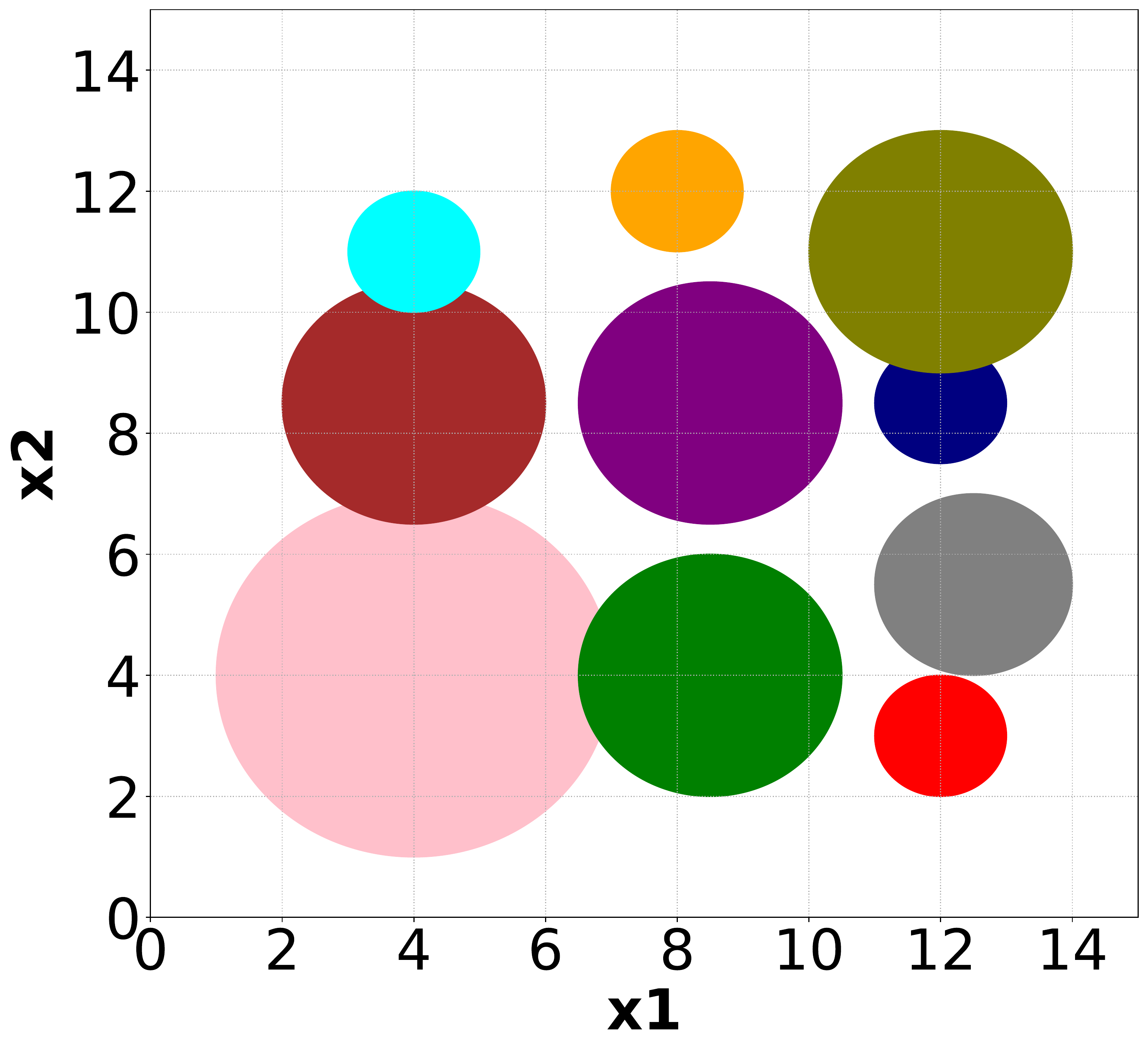}%
		\label{fig:circles10}}
	\subfloat[drifted]{\includegraphics[scale=0.12]{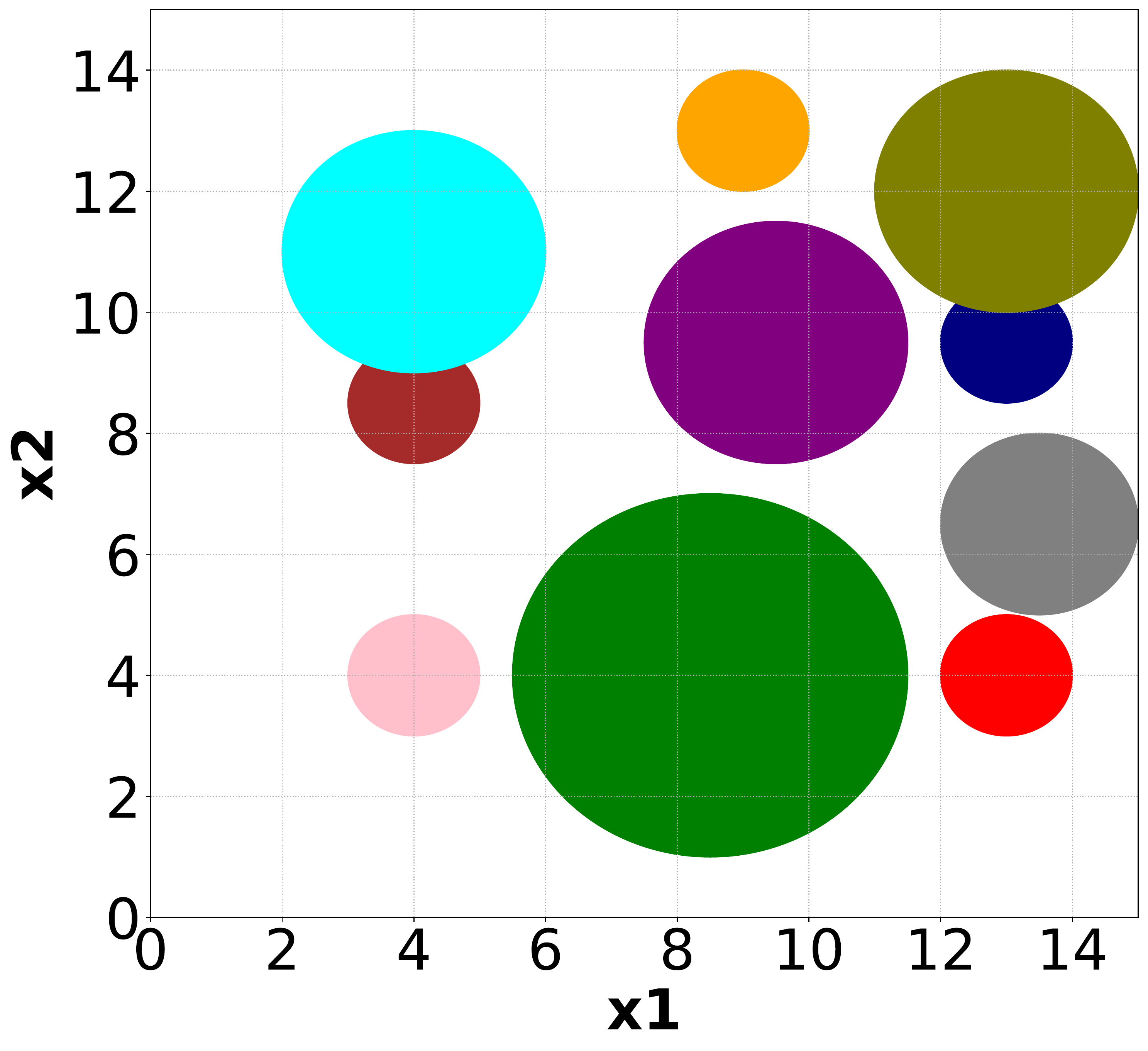}%
		\label{fig:circles10_drifted}}
	
	\caption{The \textit{circles10} dataset which consists of ten classes.}
\end{figure}

\textbf{circles10} \cite{gama2004learning}: It has two features $x_1,x_2 \in [0, 15]$ and ten classes as shown in Fig.~\ref{fig:circles10}. Each class function is a circle given by $(x_1 - x_{1c})^2 + (x_2 - x_{2c})^2 = r_c^2$ where $(x_{1c}, x_{2c})$ is its centre and $r_c$ its radius. Data normalisation is applied so that $x_1,x_2 \in [0, 1]$.

The same dataset under concept drift is presented in Fig.~\ref{fig:circles10_drifted}. The radius of the three vertical circles on the left (pink, brown, cyan) and the green circle has been changed, while all the remaining circles have been drifted by $+1$ in both dimensions. Concept drift affects all the classes \textit{simultaneously} and \textit{immediately} (abruptly) in our experiments.

Initially, the dataset is balanced i.e. the probability of an arriving instance belonging to any class is $p(y) = 0.1$. We also conducted experiments in a multi-minority scenario where the probability of an arriving instance belonging to a specific class (pink circle) is $p(y) = 0.955$, while for the rest is $p(y) = 0.005$. This constitutes a case of extreme imbalance, which creates significant problems for most learning algorithms.

Lastly, the \textit{circles10} is more challenging than \textit{sea4}, not only because it has a larger number of classes, but also because the data is noisy. We can observe from Fig.~\ref{fig:circles10} the overlap between some circles, in other words, examples with the same inputs may not have the same class label.

\subsection{Compared methods}
For fairness, all the approaches share the same base classifier, which is a fully-connected neural network of three hidden 32-neuron layers with parameters as follows: \textit{He Normal} \cite{he2015delving} weight initialisation, learning rate of $0.01$, the \textit{Rectified Adam} \cite{liu2019variance} optimisation algorithm, \textit{LeakyReLU} \cite{maas2013rectifier} as the activation function of the hidden neurons and mini-batch size of 64. Note that the classifier is only trained once per time step (i.e. $num\_epochs = 1$) as, in practise, this would allow learning in high-speed data applications. For the siamese network, the \textit{sigmoid} activation and the \textit{binary cross-entropy} loss function are used, while for a fully-connected network, the \textit{softmax} activation and the \textit{categorical cross-entropy} loss function. The following algorithms are compared in our study:

\textbf{incremental}: The state-of-the-art incremental learning algorithm \cite{zliobaite2013active} which initially proposed the active learning strategy in Eq.~\ref{eq:strategy} and the budget spending mechanism in Eq.~\ref{eq:budget}. We use the following parameters as recommended by \cite{zliobaite2013active}: step size parameter $s = 0.01$, randomisation threshold $\delta = 1.0$ and sliding window $w = 300$.

\textbf{ActiQ}: It uses the proposed learning approach but instead of a siamese network, it uses the fully-connected neural network described earlier. It is similar to the previous one but it is not incremental as it makes use of older examples in $Q$. In all experiments $E=5$.

\textbf{ActiSiamese}: This is the proposed approach as discussed in Section~\ref{sec:proposed} and its pseudocode is given in Algorithm~\ref{alg:proposed}.

To make this comparison as fair as possible, in addition to the fact that all approaches share the same active learning strategy and base classifier, we do not allow any offline learning. In other words, learning starts at time $t=0$ for all compared methods, even if the \textit{ActiQ} and \textit{ActiSiamese} have access to the initial labelled set of $E$ examples.

\subsection{Performance metrics}
Classifiers are typically evaluated using the overall accuracy metric. When class imbalance exists, however, this metric becomes problematic as it is biased towards the majority class(es) \cite{he2008learning}. Hence, it is necessary to use a metric which is not sensitive to imbalance. The geometric mean is such a metric which is defined as follows \cite{sun2006boosting}: 
\begin{equation}\label{eq:gmean}
G\text{-}mean = \displaystyle\sqrt[K]{\prod_{c=1}^K R_c},
\end{equation}
\noindent where $R_c$ is the recall for class $c$.

\subsection{Evaluation method}
To evaluate predictive sequential learning algorithms, we adopt the \textit{prequential error with fading factors} method. It has been proven that for learning algorithms in stationary data this method converges to the Bayes error \cite{gama2013evaluating}. This method does not require a holdout set and the predictive model is always tested on unseen data. We have set the fading factor to $\theta = 0.99$. In all simulations we plot the prequential \textit{G-mean} in every step averaged over 30 repetitions, including the error bars displaying the standard error around the mean.

\section{Experimental Results}\label{sec:exp_results}

\subsection{Role of the budget}
\begin{figure}[t!]
	\centering
	
	\subfloat[\textit{sea4}]{\includegraphics[scale=0.135]{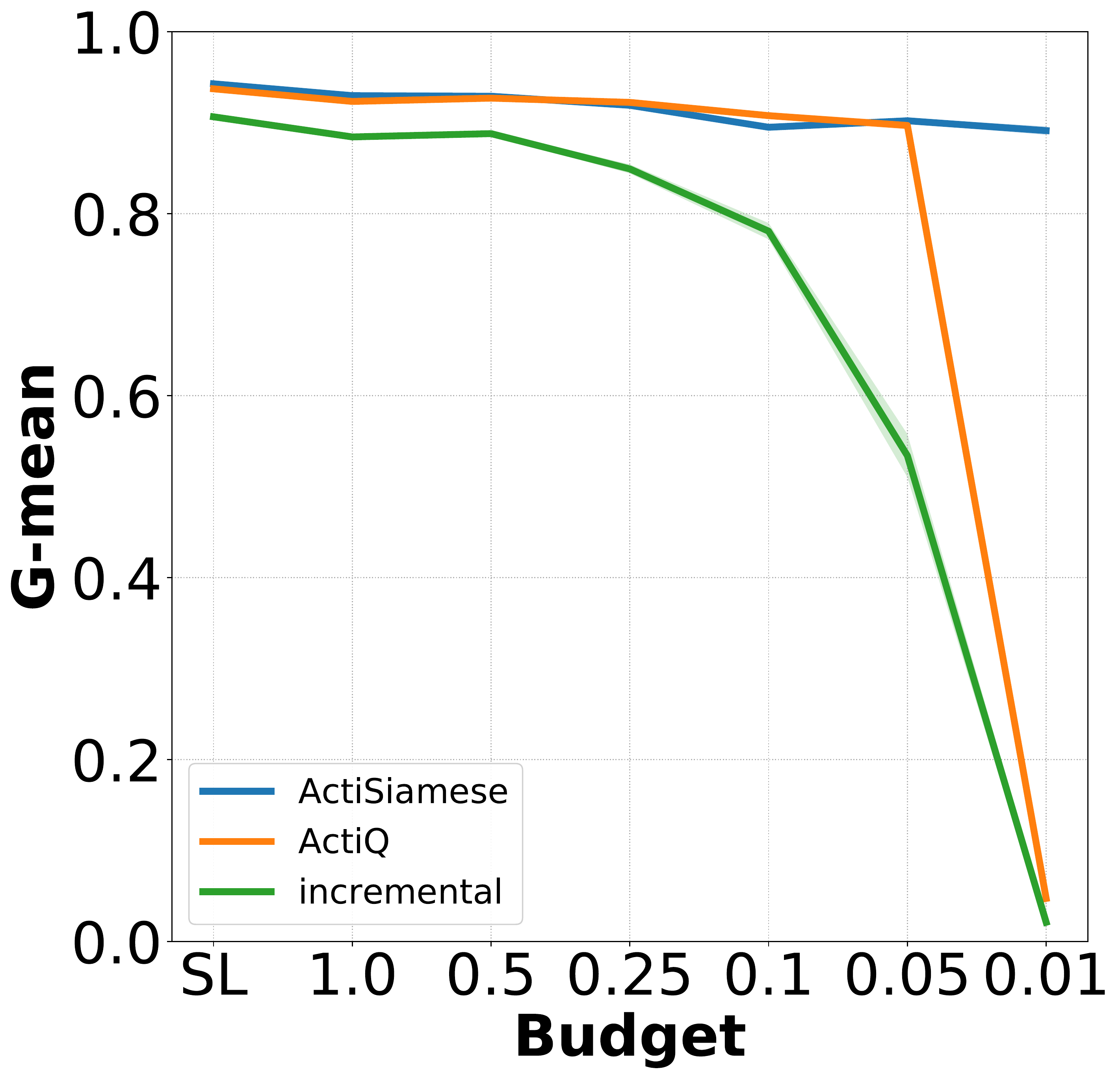}%
		\label{fig:sea4_overall}}
	\subfloat[\textit{circles10}]{\includegraphics[scale=0.135]{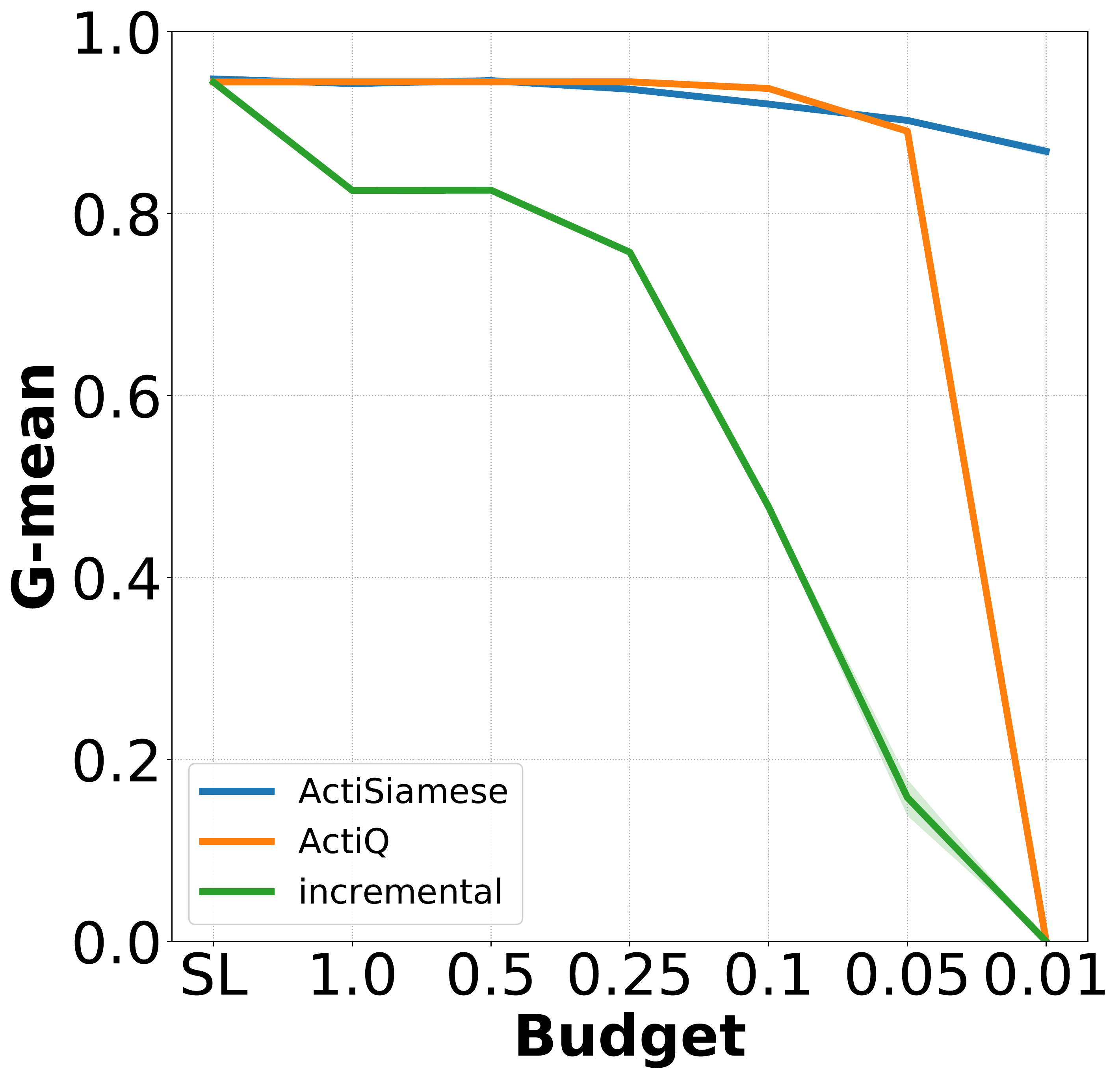}%
		\label{fig:circles10_overall}}
	
	\caption{Role of the budget in the final performance ($E = 5$)}
\end{figure}

The first experimental series examines how various budget values affect the \textit{final performance} (i.e. the prequential $G\text{-}mean$ at the final time step of a simulation). Figs~\ref{fig:sea4_overall} and \ref{fig:circles10_overall} show the results for \textit{sea4} and \textit{circles10} respectively. $SL$ refers to fully supervised online learning. The general trend is that as the budget is reduced, the final performance declines. The difference lies in how rapidly or slowly this decline occurs.

\begin{figure}[t]
	\centering
	
	\subfloat[\textit{sea4}]{\includegraphics[scale=0.135]{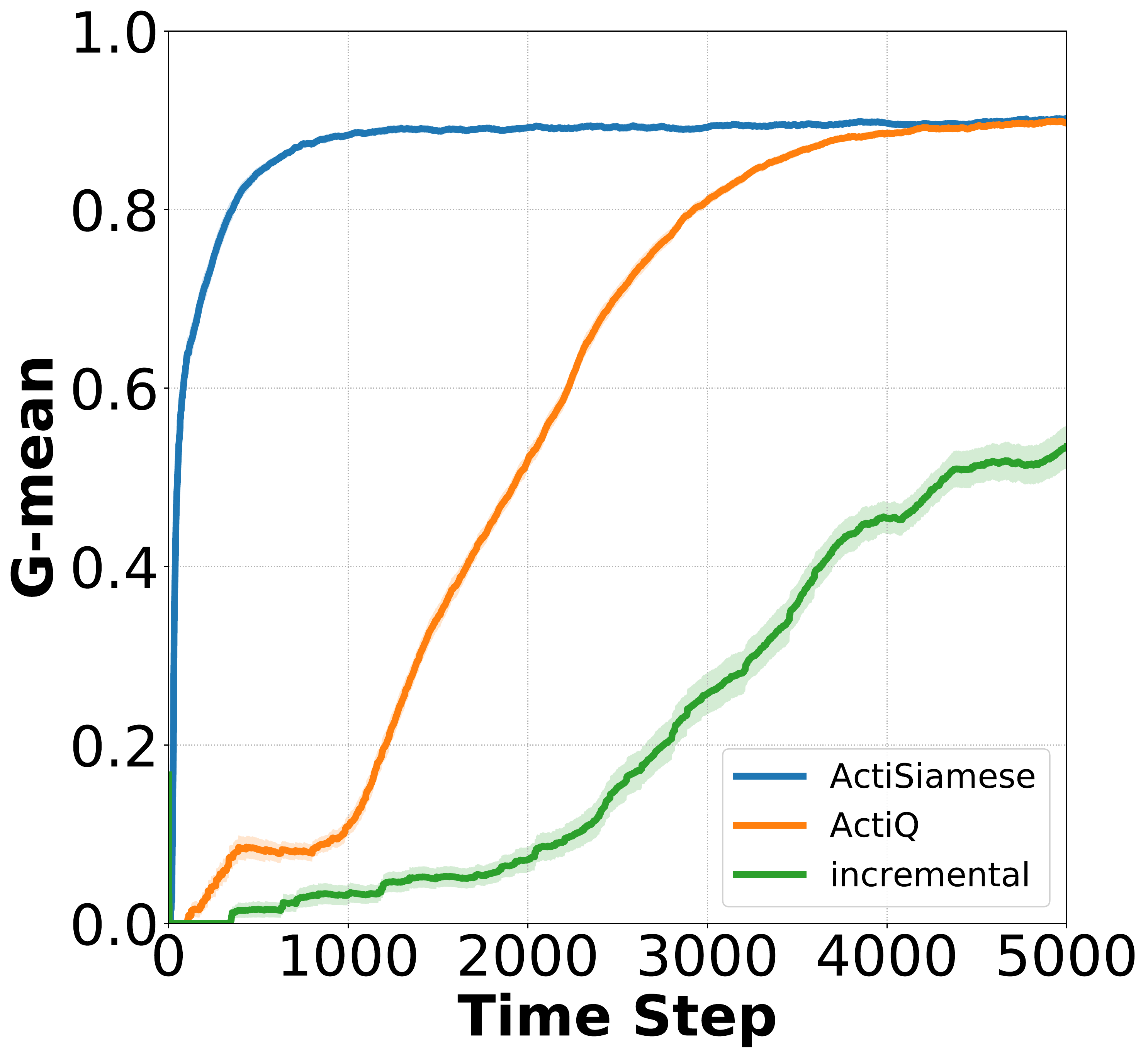}%
		\label{fig:sea4_budget005}}
	\subfloat[\textit{circles10}]{\includegraphics[scale=0.135]{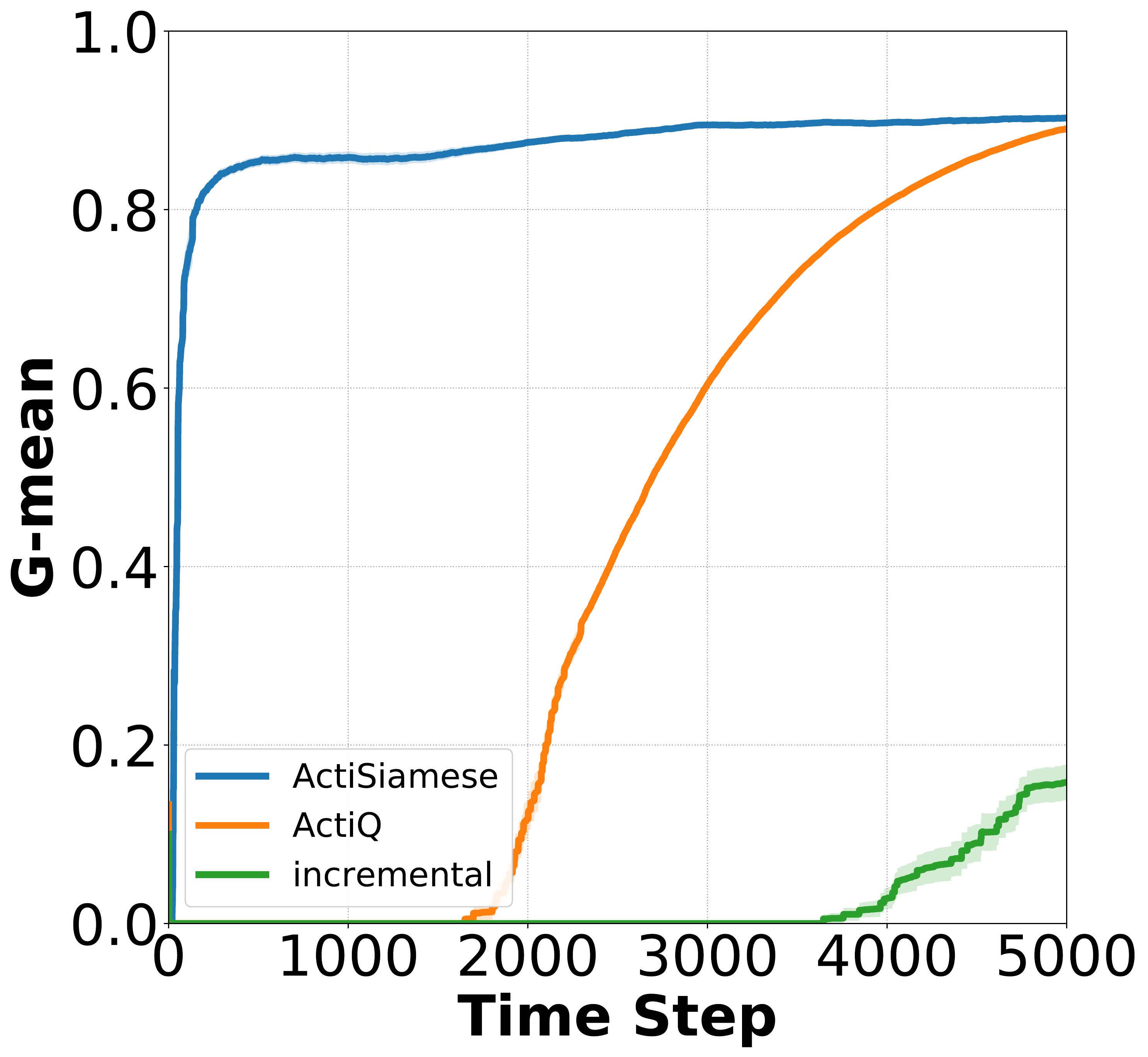}%
		\label{fig:circles10_budget005}}
	
	\caption{Comparative study in stationary settings ($B=0.05$)}
\end{figure}

When the budget is $B=0.01$, the \textit{ActiQ} and \textit{incremental} obtain a score of $G\text{-}mean = 0$. The \textit{ActiSiamese} approach significantly outperforms the rest. In fact, by having access to $1\%$ of the labels, the proposed approach sacrifices only $0.5\%$ of its performance. We consider this to be a significant advantage of the proposed approach as it can potentially enable the realistic deployment of an online classifier.

For greater values of budget ($B > 0.01$) the \textit{ActiSiamese} and \textit{ActiQ} obtain almost identical final performance scores. As we will discuss in the next section, however, their learning speed is significantly different. The \textit{incremental} approach almost always performs significantly worse. We do acknowledge, of course, that this approach does not store or use any older examples, contrary to the other two.

\subsection{Stationary data}
The previous experiments only consider the algorithms' final performance. This section examines another important characteristic, that is, their \textit{learning speed}. We focus on the interesting case of $B=0.05$ as \textit{ActiQ} and \textit{ActiSiamese} appear to achieve a similar final performance. Figs.~\ref{fig:sea4_budget005} and \ref{fig:circles10_budget005} show a comparison for $B=0.05$ in the \textit{sea4} and \textit{circles10} dataset respectively. Despite the fact that the two approaches obtain a similar final G-mean score, it can be observed that the \textit{ActiQ} requires about 4000 and 5000 time steps respectively to equalise the \textit{ActiSiamese}'s performance. This is another major advantage of the proposed learning approach since in \textit{online} learning, speed is a crucial performance measure. If a proposed solution is slow, it may be practically useless even if it eventually achieves the correct result.

\begin{figure}[t!]
	\centering
	
	\subfloat[\textit{sea4}]{\includegraphics[scale=0.135]{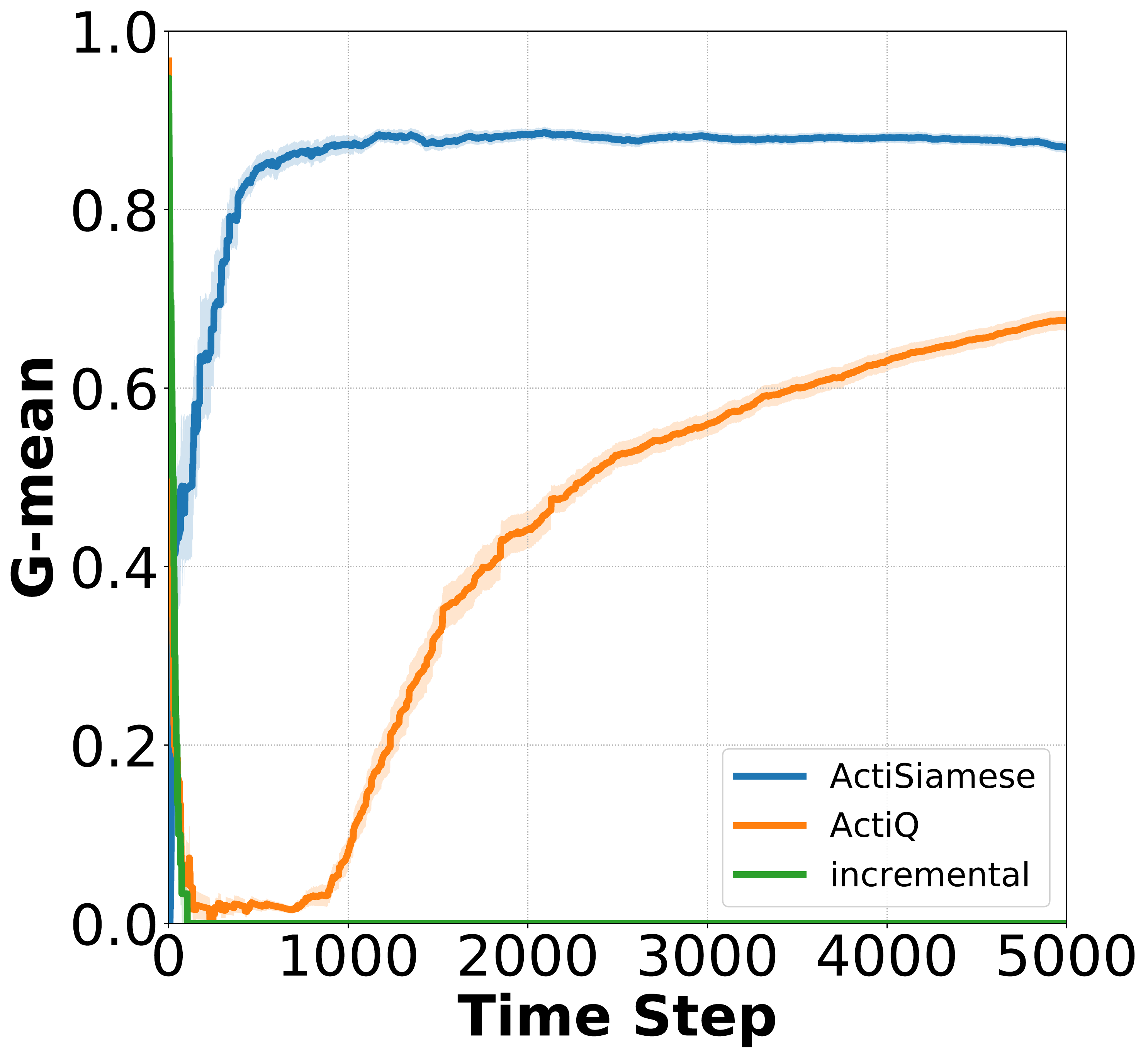}%
		\label{fig:sea4_budget005_mm}}
	\subfloat[\textit{circles10}]{\includegraphics[scale=0.135]{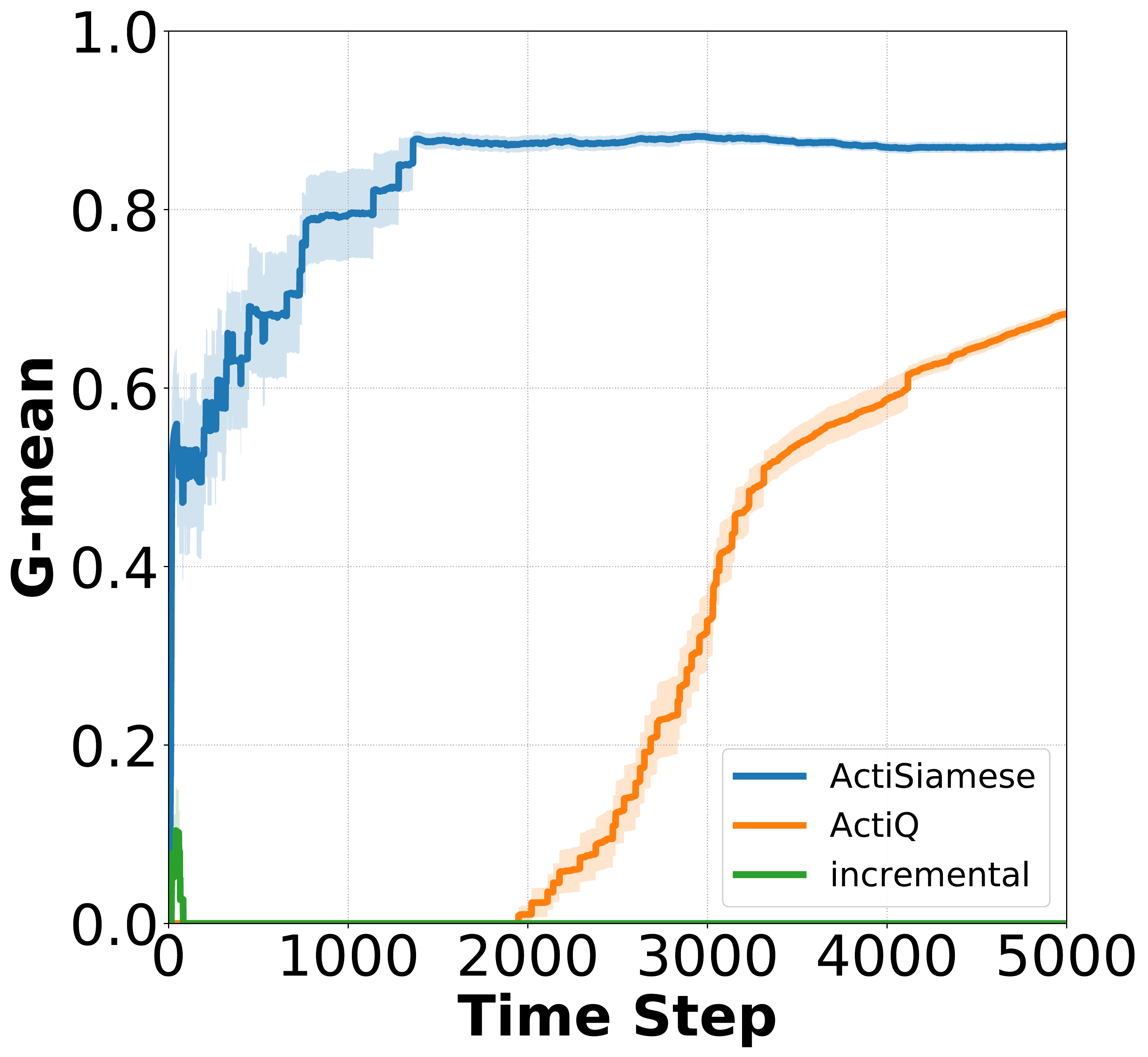}%
		\label{fig:circles_budget005_mm}}
	
	\caption{Comparative study in imbalanced settings ($B=0.05$)}
\end{figure}

\begin{figure}[t]
	\centering
	
	\subfloat[\textit{sea4}]{\includegraphics[scale=0.135]{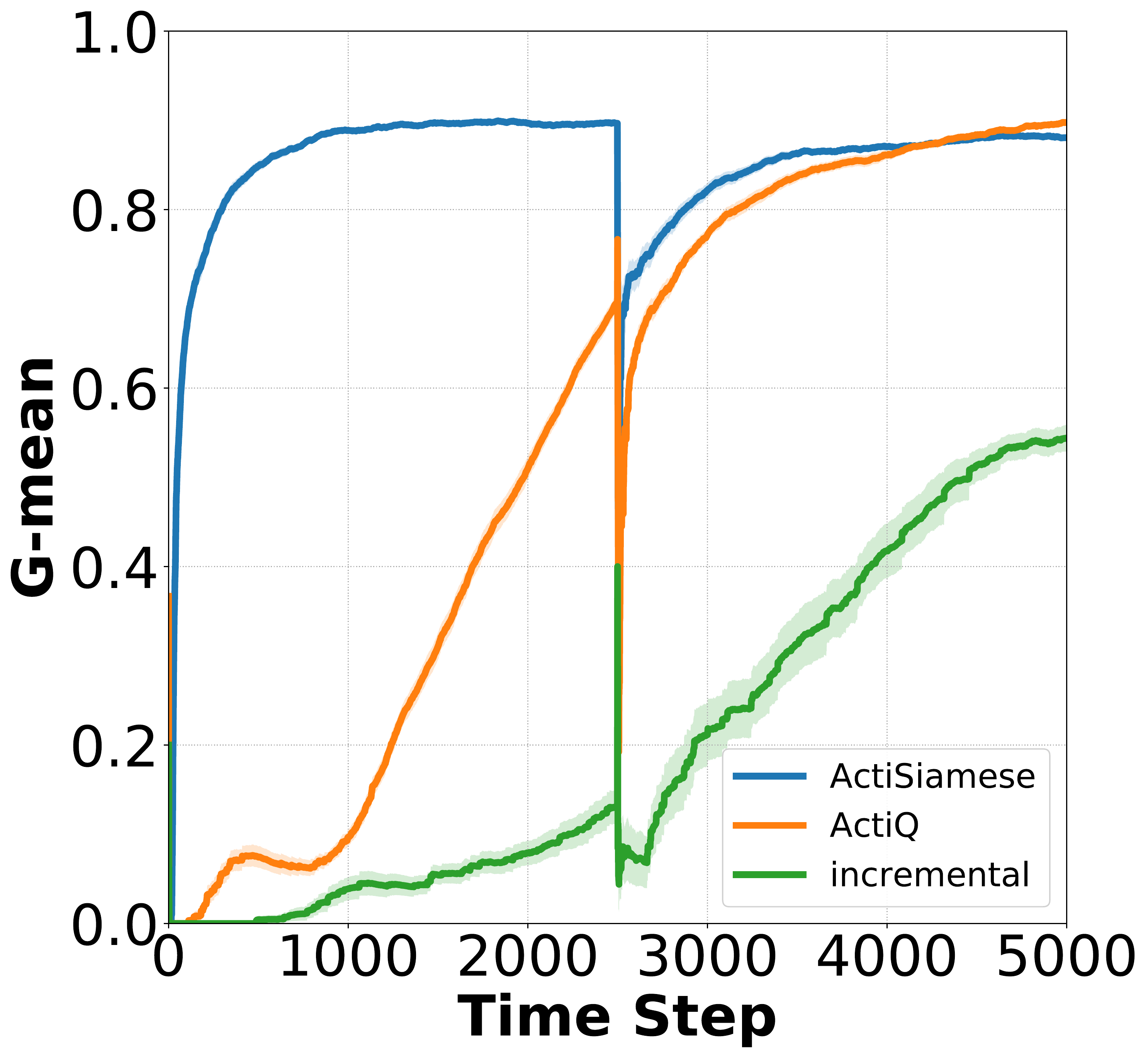}%
		\label{fig:sea4_budget005_drift}}
	\subfloat[\textit{circles10}]{\includegraphics[scale=0.135]{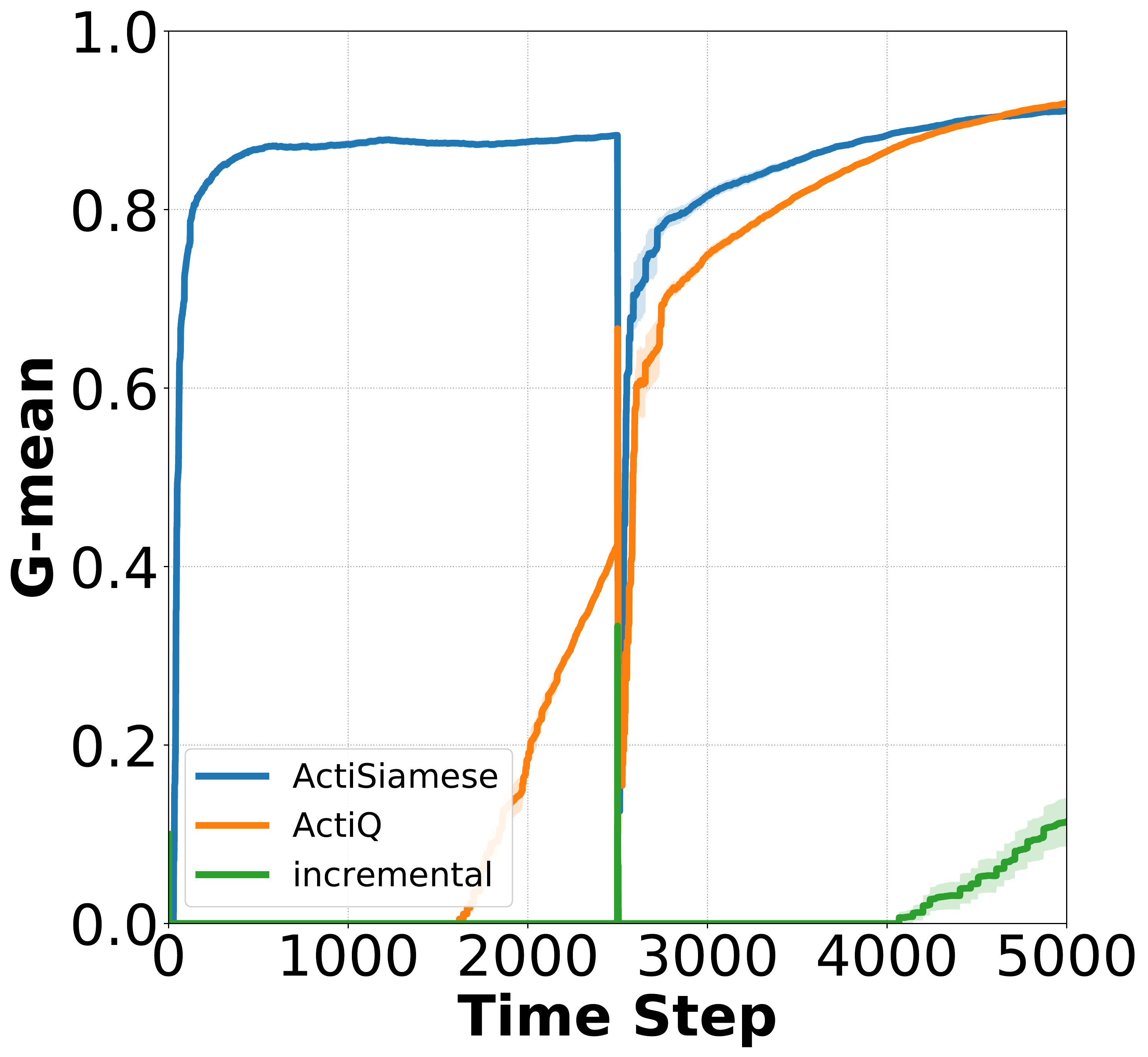}%
		\label{fig:circles10_budget005_drift}}
	
	\caption{Comparative study in nonstationary settings ($B=0.05$)}
\end{figure}

\subsection{Imbalanced data}
Fig.~\ref{fig:sea4_budget005_mm} depicts the learning curves for imbalanced data for \textit{sea4}. The \textit{ActiSiamese} approach significantly outperforms the rest as it is robust to imbalance.  The \textit{ActiQ}'s performance is severely affected by the imbalance. The \textit{incremental} obtains a score of $G\text{-}mean = 0$ as it fails to do well on the minority classes, and hence it is not visible in the figure. This is similar for \textit{circles10} in Fig.~\ref{fig:circles_budget005_mm} where the \textit{ActiSiamese} is only hindered until about $t=1300$. In both figures, the \textit{ActiQ}'s final performance is significantly worse even after 5000 time steps, something that wasn't the case in Figs.~\ref{fig:sea4_budget005} and \ref{fig:circles10_budget005}.

\subsection{Nonstationary data}
Fig.~\ref{fig:sea4_budget005_drift} depicts the performance in nonstationary data for \textit{sea4}, specifically, when drift occurs abruptly at $t=2500$. The \textit{ActiSiamese} approach is unable to fully recover, however, it does fully recover in Fig.~\ref{fig:circles10_budget005_drift} in the \textit{circles10} dataset. Interestingly, the proposed \textit{ActiQ} approach slightly outperforms the \textit{ActiSiamese} by time $t=5000$. This preliminary study reveals that there may be evidence to suggest that \textit{ActiSiamese} has a more difficult time to `forget' old concepts than \textit{ActiQ}. 

\subsection{Imbalanced and nonstationary data}
\begin{figure}[t!]
	\centering
	
	\subfloat[\textit{sea4}]{\includegraphics[scale=0.135]{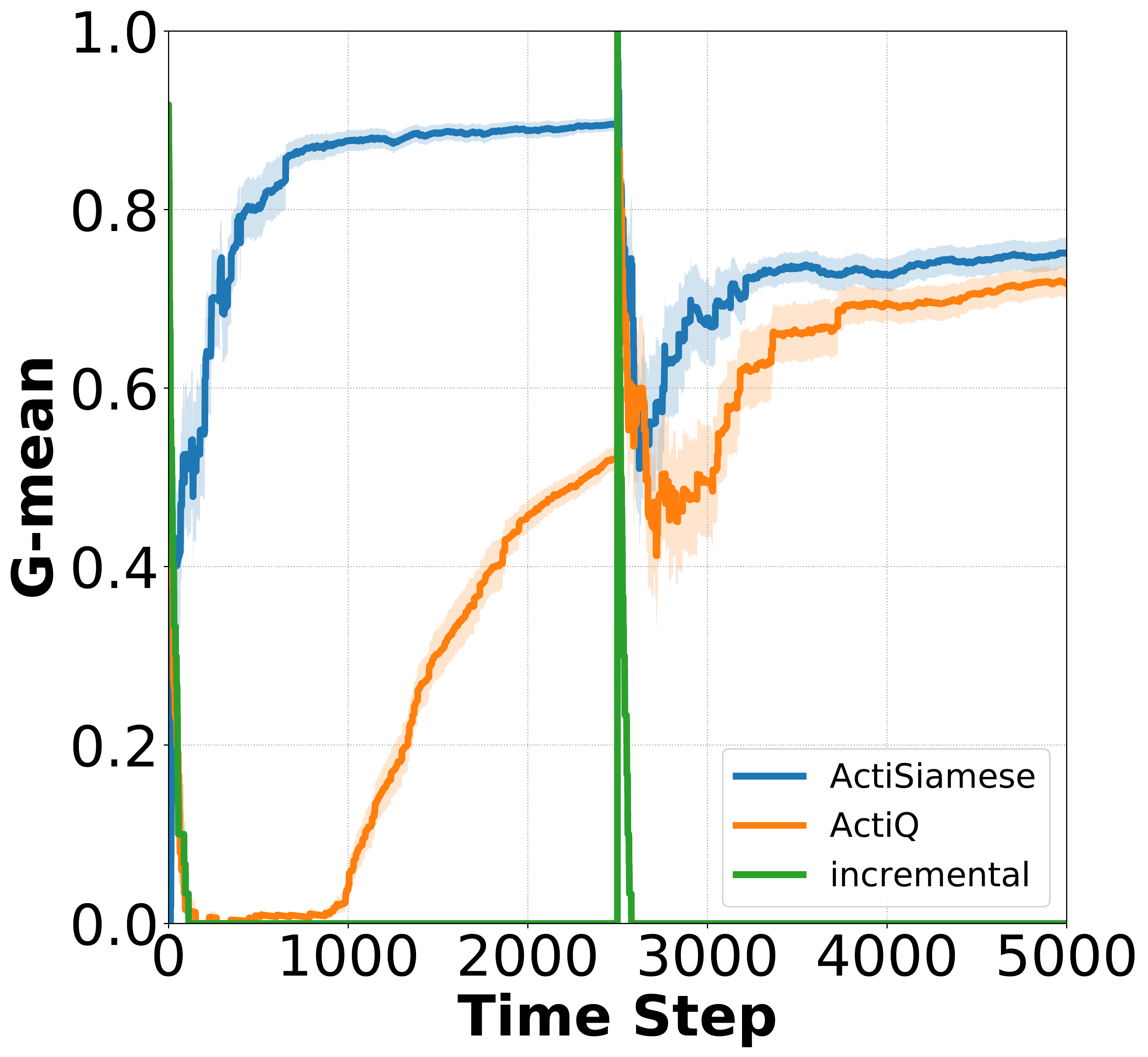}%
		\label{fig:sea4_budget005_drift_mm}}
	\subfloat[\textit{circles10}]{\includegraphics[scale=0.135]{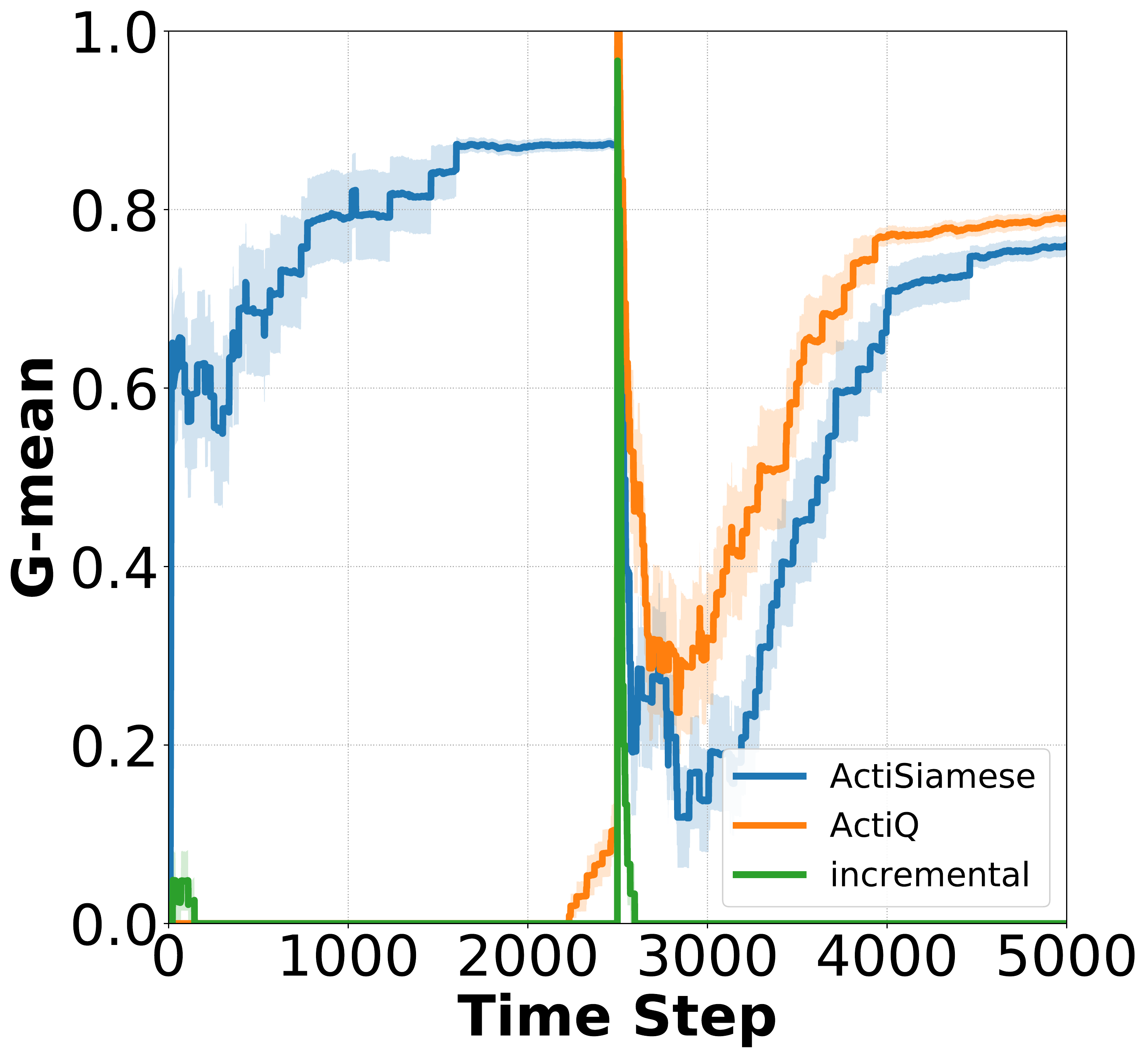}%
		\label{fig:circles10_budget005_drift_mm}}
	
	\caption{Comparative study in imbalanced \& nonstationary settings ($B=0.05$)}
\end{figure}

Fig.~\ref{fig:sea4_budget005_drift_mm} depicts the performance when data is both imbalanced and nonstationary for the \textit{sea4} dataset. After the concept drift, the \textit{ActiSiamese} approach cannot fully recover from it but performs better than \textit{ActiQ}. In Fig.~\ref{fig:circles10_budget005_drift_mm}, the \textit{ActiQ} performs better than the \textit{ActiSiamese} after the drift. The \textit{ActiSiamese}'s poor performance under these conditions is attributed to its inability to fully recover from the drift, thus reinforcing our previous finding that \textit{ActiSiamese} may have a more difficult time to `forget' old concepts. Moreover, these results indicate that when imbalance and drift co-exist and are both severe, this still remains an open and challenging problem.

\section{Conclusion and Future Work}\label{sec:remarks}
We have proposed an online learning approach that combines active learning and siamese networks to address the challenges of limited labelled, nonstationary and imbalanced data. The proposed approach significantly outperforms strong baselines and state-of-the-art algorithms in terms of both learning speed and performance and has been shown to be effective even when only $1\%$ of the labels of the arriving instances is available, something which is not unrealistic in deployed settings. For future work, we will enrich our study with real-world datasets. The problem of learning from nonstationary and imbalanced data still remains open. We have shown that when imbalance and drift co-exist and are both severe, all compared algorithms are severely affected. We plan to make the proposed algorithms more robust to these conditions.

\section*{Acknowledgment}
This work has been supported by the European Union's Horizon 2020 research and innovation programme under grant agreements No. 867433 (FAULT-LEARNING) and No. 739551 (KIOS CoE), and from the Republic of Cyprus through the Directorate General for European Programmes, Coordination and Development.

\bibliographystyle{IEEEtran}
\bibliography{ijcnn20}

\end{document}